\documentclass[11pt]{article}

\usepackage[margin=1in]{geometry}
\usepackage{amsmath,amssymb,amsfonts}
\usepackage{graphicx}
\usepackage{booktabs}
\usepackage{hyperref}
\usepackage{cleveref}
\usepackage{natbib}
\usepackage{xcolor}
\usepackage{caption}
\usepackage{subcaption}
\usepackage{microtype}
\usepackage{enumitem}
\usepackage{multirow}
\usepackage{tcolorbox}
\usepackage{wrapfig}

\graphicspath{{figures/}}

\newcommand{\R}{\mathbb{R}}
\newcommand{\norm}[1]{\left\|#1\right\|}

\title{The Lifecycle of the Spectral Edge:\\From Gradient Learning to Weight-Decay Compression}

\author{%
  Yongzhong Xu%
  \thanks{abbyxu@gmail.com; code at \url{https://github.com/skydancerosel/dyck\_scan}}
}

\date{}

\begin{document}
\maketitle

% ══════════════════════════════════════════════════════════════════════
\begin{abstract}
% ══════════════════════════════════════════════════════════════════════

The spectral edge---the dominant direction of the Gram matrix of parameter updates---has been shown to track phase transitions in neural network training.
But what \emph{drives} the spectral edge, and why does it matter causally?

We decompose the spectral edge into its gradient and weight-decay components across two sequence tasks (Dyck-1 balanced parentheses and SCAN compositional generalization) and discover a sharp two-phase lifecycle.
\textbf{Before grokking}, the edge is gradient-driven (88--98\% gradient energy) and carries task-relevant functional content.
\textbf{At grokking}, gradient and weight decay \emph{align} along the edge---both push the same direction---and the edge transitions to a compression mode (0.2--5\% gradient, 95--99.8\% weight decay).
This alignment is the microscopic signature of the phase transition.

The post-grok edge presents a paradox: perturbing along it changes almost nothing (max KL $\approx 0.00005$; Hessian curvature $0.08$), yet removing it is catastrophic ($\Delta\text{acc} = -0.58 \pm 0.09$; $>$4000$\times$ worse than removing random directions).
The resolution is geometric: the model's function depends on the edge's \emph{orientation} (locked by the spectral gap) but not on displacement \emph{along} it (flat curvature).
The edge defines a stable axis that weight decay compresses without disrupting.

Three universality classes emerge:
\emph{functional edges} (modular arithmetic: edge carries distinct Fourier modes),
\emph{mixed edges} (Dyck: edge retains partial functional content),
and \emph{compression edges} (SCAN: edge is functionally empty, frozen at 8$^\circ$ rotation).
The class is predicted by the balance of gradient driving vs.\ weight-decay damping in the gap flow equation.

Six causal experiments establish the mechanism: grad-WD decomposition, ablation with random controls, $\varepsilon$-sweep perturbation curves, Hessian curvature, nonlinear probes (MLP $R^2 = 0.99$ where linear $R^2 = 0.86$), and a weight-decay intervention showing that removing WD post-grok reverses compression ($R^2_\text{linear}: 0.85 \to 0.99$) while preserving the learned algorithm (accuracy 0.97).
All findings replicate across 3 seeds.

\end{abstract}

% ══════════════════════════════════════════════════════════════════════
\section{Introduction}
\label{sec:intro}
% ══════════════════════════════════════════════════════════════════════

Training dynamics of neural networks are highly structured despite the enormous dimensionality of parameter space.
The spectral edge thesis~\citep{xu2026spectral_edge} makes this precise: the Gram matrix of rolling-window parameter updates develops an intra-signal eigenvalue gap that separates a few dominant directions from the bulk.
These dominant directions---the spectral edge---concentrate the variance of training, and phase transitions in learning coincide with gap events~\citep{xu2026functional_modes, xu2026lowdim, xu2026commutator}.

The thesis establishes \emph{that} the spectral edge exists and tracks learning.
This paper asks \emph{why}: what drives the spectral edge, and what gives it causal importance?

\subsection{The Central Claim}

\begin{tcolorbox}[colback=blue!3, colframe=blue!40, boxrule=0.5pt, arc=2pt]
\textbf{Claim (The Two-Phase Lifecycle).}
During training with weight decay, the dominant singular vector $v_1$ of the Gram matrix of parameter updates (defined in \S\ref{sec:setup}) undergoes a sharp transition:
\begin{enumerate}[nosep]
    \item \textbf{Learning phase}: $v_1$ is gradient-driven, functionally active, and carries task-relevant information.
    \item \textbf{Transition}: at grokking, gradient and weight decay \emph{align} along $v_1$ (both push the same direction).
    \item \textbf{Compression phase}: $v_1$ becomes weight-decay-driven, functionally flat, but causally essential---the model's function depends on the edge's orientation but is invariant to displacement along it.
\end{enumerate}
\end{tcolorbox}

\noindent
Everything in this paper supports this claim through multiple projections of the same underlying mechanism:
\begin{itemize}[nosep]
    \item In \emph{weight space}: the grad-WD decomposition (\S\ref{sec:mechanism}).
    \item In \emph{function space}: the ablation paradox and perturbation flatness (\S\ref{sec:paradox}).
    \item In \emph{representation space}: nonlinear re-encoding and the probe $R^2$ inversion (\S\ref{sec:nonlinear}).
    \item In the \emph{gap flow equation}: the Term~2/Term~3 balance predicts universality classes (\S\ref{sec:universality}).
\end{itemize}

\subsection{Relation to Prior Work}

The spectral edge thesis~\citep{xu2026spectral_edge} derives gap dynamics from three axioms and confirms 19/20 quantitative predictions across six model families spanning 150K to 124M parameters (TinyStories 51M, GPT-2 124M, and grokking experiments on Dyck, SCAN, and modular arithmetic).
A companion paper~\citep{xu2026functional_modes} shows that edge directions are \emph{functional modes}---structured perturbation patterns that collapse to Fourier frequencies for modular arithmetic.
The commutator analysis~\citep{xu2026commutator, xu2026transverse} tracks $\norm{[W_Q, W_K]}_F$ and shows it peaks before grokking, with superlinear lead times across Dyck, SCAN, and modular arithmetic.

All three works describe the spectral edge as a \emph{geometric object}: eigenvalue gaps, singular vectors, commutator norms.
This paper adds a \emph{dynamical decomposition}: the edge is not one thing---it transitions from a gradient-driven learning direction to a weight-decay-driven compression axis at the moment of grokking.
This resolves the open question of why the spectral gap continues to widen post-grok even though learning has already occurred: the post-grok gap is maintained by weight decay compression, not by ongoing learning.

% ══════════════════════════════════════════════════════════════════════
\section{Setup}
\label{sec:setup}
% ══════════════════════════════════════════════════════════════════════

\subsection{Tasks}

\paragraph{Dyck-1 depth prediction.}
A 2-layer causal Transformer ($d_\text{model} = 128$, 4 heads, ${\sim}$150K params) predicts stack depth (0--12) at each position in balanced parentheses sequences.
Training: 50 sequences; test: 5000.
The computation is a cumulative sum: $\text{depth}(t) = \sum_{i \leq t} s_i$ where $s_i = +1$ (open) or $-1$ (close).

\paragraph{SCAN.}
A 6-layer encoder--decoder Transformer ($d_\text{model} = 256$, 4 heads, ${\sim}$1.5M params) translates commands (``jump left twice'') to actions.
Training: 2048 pairs; test: 500.
The computation factors compositionally: verb $\times$ direction $\times$ repetition.

\paragraph{Grokking.}
Both tasks are trained with AdamW ($\beta_2 = 0.98$), weight decay $\omega \in \{0, 1\}$, 3 seeds each.
With $\omega = 1$: Dyck groks at steps 600--1400, SCAN at 2500--4000.
With $\omega = 0$: no grokking in any run.
Hit rate: 6/6 grok with WD, 0/6 without, per task.
Combined with the thesis's modular arithmetic runs~\citep{xu2026spectral_edge}, this gives 24/24 grokking with WD and 0/24 without, across four task families.

\subsection{Spectral Edge Construction}

Following the thesis, we compute the trajectory matrix $X(t) \in \R^{W \times p}$ from $W = 5$ consecutive parameter update deltas (restricted to attention weights), extract singular values $\sigma_1 \geq \cdots \geq \sigma_W$ and right singular vectors $v_1, \ldots, v_W \in \R^p$.
The spectral gap $g_{23} = \sigma_2^2 - \sigma_3^2$ compresses $33\times$ (Dyck) and $43\times$ (SCAN) during grokking, with $k^* = 1$ universally---replicating the thesis.

% ══════════════════════════════════════════════════════════════════════
\section{The Mechanism: Gradient--Weight-Decay Alignment}
\label{sec:mechanism}
% ══════════════════════════════════════════════════════════════════════

\subsection{Decomposition}

AdamW's parameter update decomposes exactly:
\[
    \Delta\theta = \underbrace{\Delta\theta_\text{grad}}_{\text{Adam-processed gradient}} + \underbrace{\Delta\theta_\text{wd}}_{ = \;-\eta\omega\theta}.
\]
We project each component onto the Gram singular vectors $v_k$ and measure the fraction of update energy from each source.

\begin{figure}[t]
    \centering
    \includegraphics[width=\textwidth]{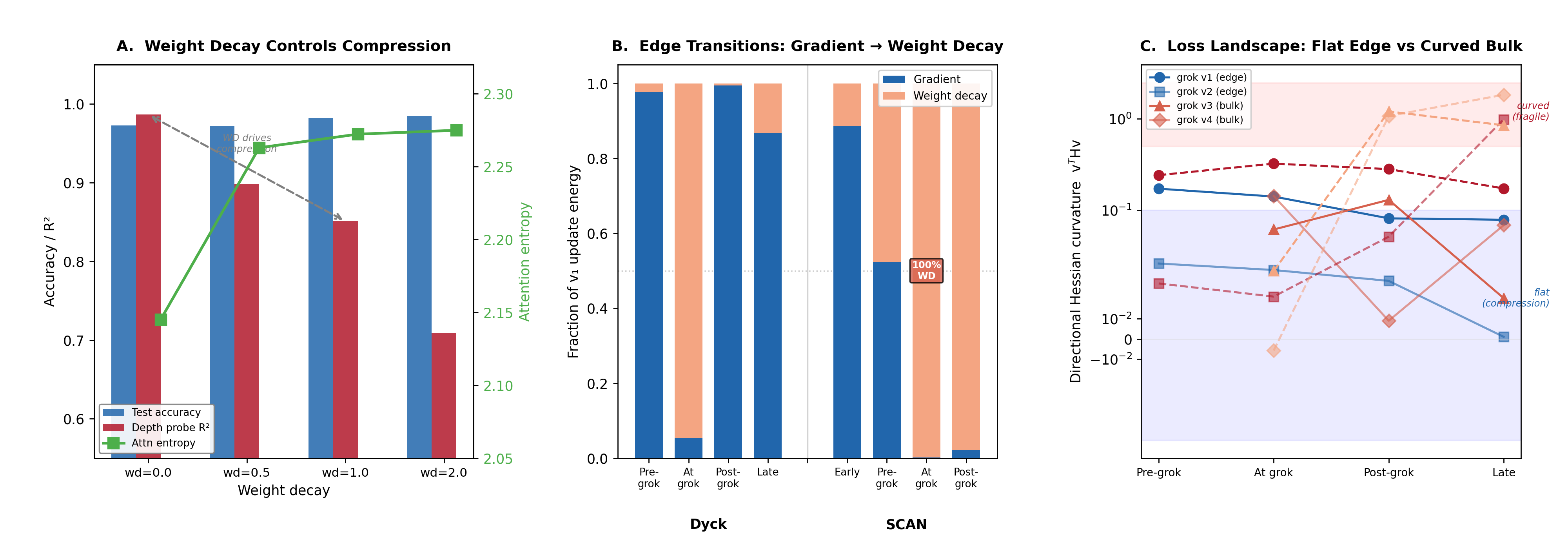}
    \caption{\textbf{(A)}~Weight decay controls the compression--accessibility tradeoff: increasing $\omega$ raises accuracy and attention entropy but lowers linear probe $R^2$ (representations become more abstract).
    \textbf{(B)}~The gradient-to-WD transition on $v_1$: both tasks flip from gradient-dominated (blue) to WD-dominated (orange) at grokking.  SCAN's transition is sharper (99.8\% WD).
    \textbf{(C)}~Hessian curvature: the grokked edge (solid) is flat (blue shading); the memorized bulk (dashed) is sharply curved (red shading, up to $v^\top\!Hv = 1.84$).}
    \label{fig:mechanism}
\end{figure}

\subsection{The Flip}

\Cref{fig:mechanism}B shows the main result.
Before grokking, $v_1$ is gradient-driven: 97.6\% (Dyck) and 88.7\% (SCAN) of the update energy along $v_1$ comes from the gradient component.
At grokking, this flips: gradient drops to 5.3\% (Dyck) and 0.2\% (SCAN), with weight decay providing the remainder.

Crucially, gradient and weight decay are \emph{aligned} along $v_1$---they push the same direction.
On bulk directions ($v_3$, $v_4$), they \emph{oppose} (the normal dynamics: gradient descends, WD regularizes).
The alignment on $v_1$ is the phase transition signature: at grokking, the optimizer's gradient and its regularizer \emph{agree} on the edge direction.

\subsection{Why Alignment Matters}

In the thesis's gap flow equation~\citep{xu2026spectral_edge}:
\begin{equation}
\frac{dg}{dt} \approx \underbrace{-\eta(h_{k^*} - h_{k^*+1}) \bar{d}}_{\text{Term 1: curvature}} - \underbrace{\eta(\bar{h} + \omega) g}_{\text{Term 2: WD damping}} + \underbrace{\eta W \!\left(\frac{|G_{k^*}|^2}{d_{k^*}} - \frac{|G_{k^*+1}|^2}{d_{k^*+1}}\right)}_{\text{Term 3: gradient driving}},
\label{eq:gap_flow}
\end{equation}
our decomposition directly measures the balance of Terms~2 and~3.
Pre-grok: Term~3 dominates (gradient drives the gap open).
At grok: Term~2 takes over (WD maintains the gap).
The grad-WD alignment is the moment when the dominant force shifts from gradient to regularizer---the edge stops \emph{learning} and starts \emph{compressing}.

\subsection{Correlational Evidence: No Alignment Without Grokking}

In all 6 control runs ($\omega = 0$), there is no weight decay and hence no WD component to align with.
Correspondingly, no grokking occurs.
Across all 12 runs (6 grok, 6 control) per task, the correspondence is perfect: alignment appears if and only if grokking occurs.
This is correlational, not causal---alignment could be a consequence rather than a cause of generalization---but the WD intervention (\S\ref{sec:wd_intervention}) provides the causal direction: WD is necessary for alignment, and alignment is necessary for the edge to transition from learning to compression.

% ══════════════════════════════════════════════════════════════════════
\section{The Ablation Paradox: Flat but Essential}
\label{sec:paradox}
% ══════════════════════════════════════════════════════════════════════

The post-grok edge presents an apparent contradiction.

\subsection{The Edge Is Uniquely Important}

\begin{figure}[t]
    \centering
    \includegraphics[width=0.85\textwidth]{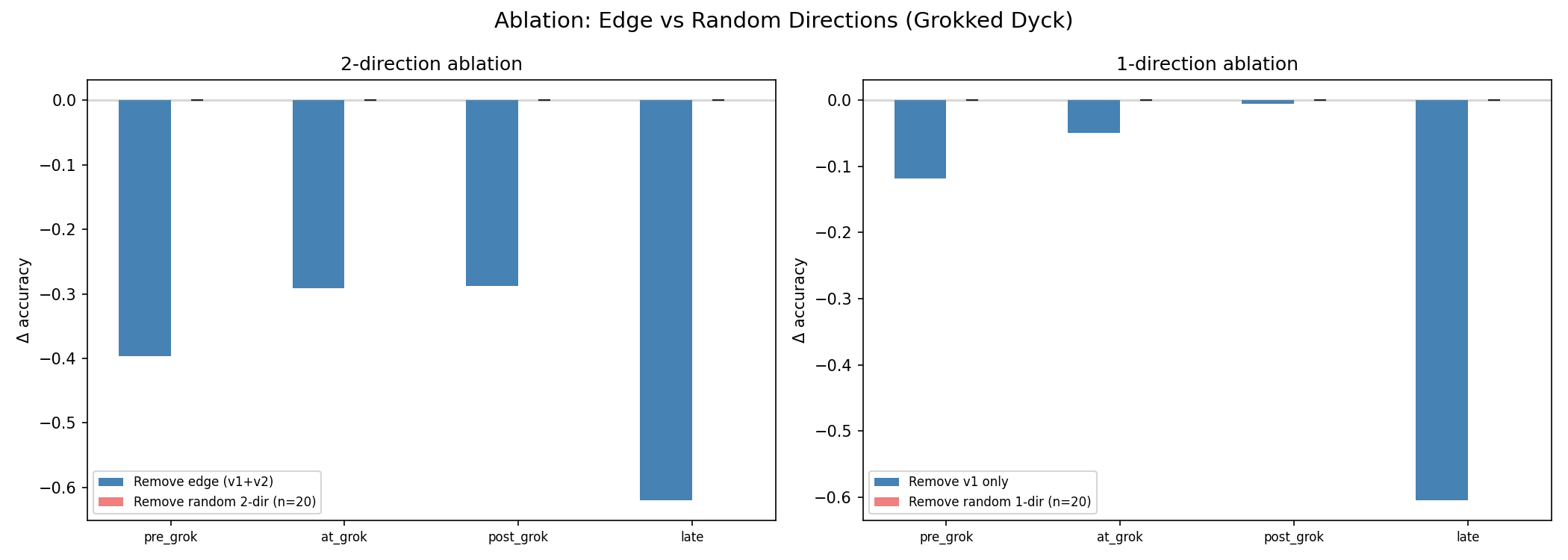}
    \caption{Ablation: removing the Gram edge ($v_1 + v_2$, blue) degrades accuracy by 0.29--0.62; removing random 2-dimensional subspaces (red, 20 trials/phase) has zero effect ($\Delta\text{acc} < 10^{-4}$).
    The edge is $>$4000$\times$ more impactful than random directions.  This is not a norm effect---the random directions remove the same number of dimensions from the same parameter subspace.}
    \label{fig:ablation}
\end{figure}

Projecting out $v_1 + v_2$ from the model's attention weights degrades accuracy by 0.26--0.62 across training phases (\Cref{fig:ablation}).
Projecting out random 2-dimensional subspaces has literally zero effect---20 trials per phase, all $|\Delta\text{acc}| < 10^{-4}$.
The edge is $>$4000$\times$ more impactful than random.
This holds for both tasks: SCAN edge ablation gives $\Delta\text{acc} = -0.49$ for the grokked model vs.\ $-0.03$ for the memorized model (where the edge has collapsed to near-zero magnitude).

\subsection{Yet the Edge Is Functionally Flat}

\begin{figure}[t]
    \centering
    \includegraphics[width=\textwidth]{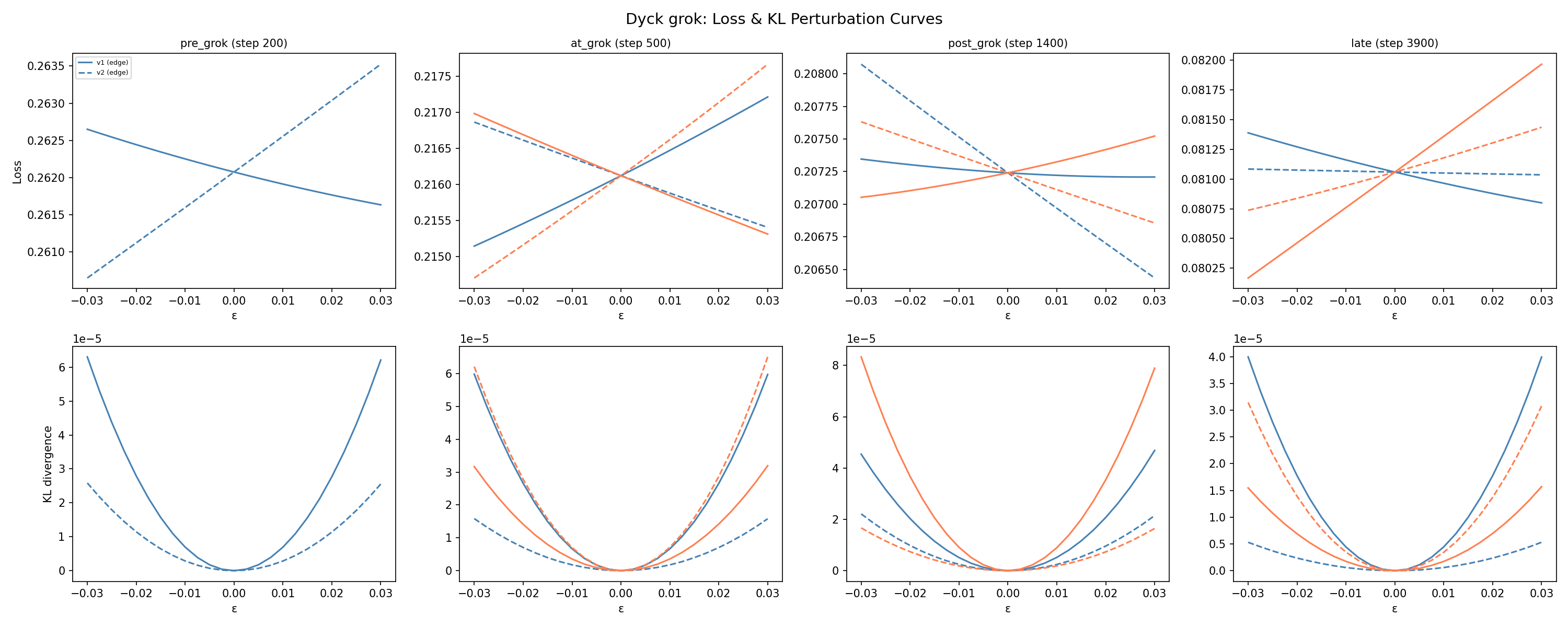}
    \caption{Perturbation curves ($\varepsilon$-sweep) for the grokked Dyck model.
    Top: loss; bottom: KL divergence from base model.
    All curves are nearly flat---edge and bulk alike.
    The grokked model sits in a wide valley where perturbation along \emph{any} Gram direction barely changes the output.
    Maximum KL $\approx 0.00005$ for the edge.}
    \label{fig:eps_curves}
\end{figure}

Perturbing the model \emph{along} $v_1$ (rather than removing it) barely changes the output (\Cref{fig:eps_curves}).
Max KL divergence $\approx 0.00005$; Hessian curvature $v_1^\top\! H v_1 = 0.078$ (\Cref{fig:mechanism}C).
The entire Gram subspace is flat for the grokked model.
The memorized model's bulk, by contrast, reaches curvature 1.84---a narrow, fragile minimum.

\begin{figure}[t]
    \centering
    \includegraphics[width=0.7\textwidth]{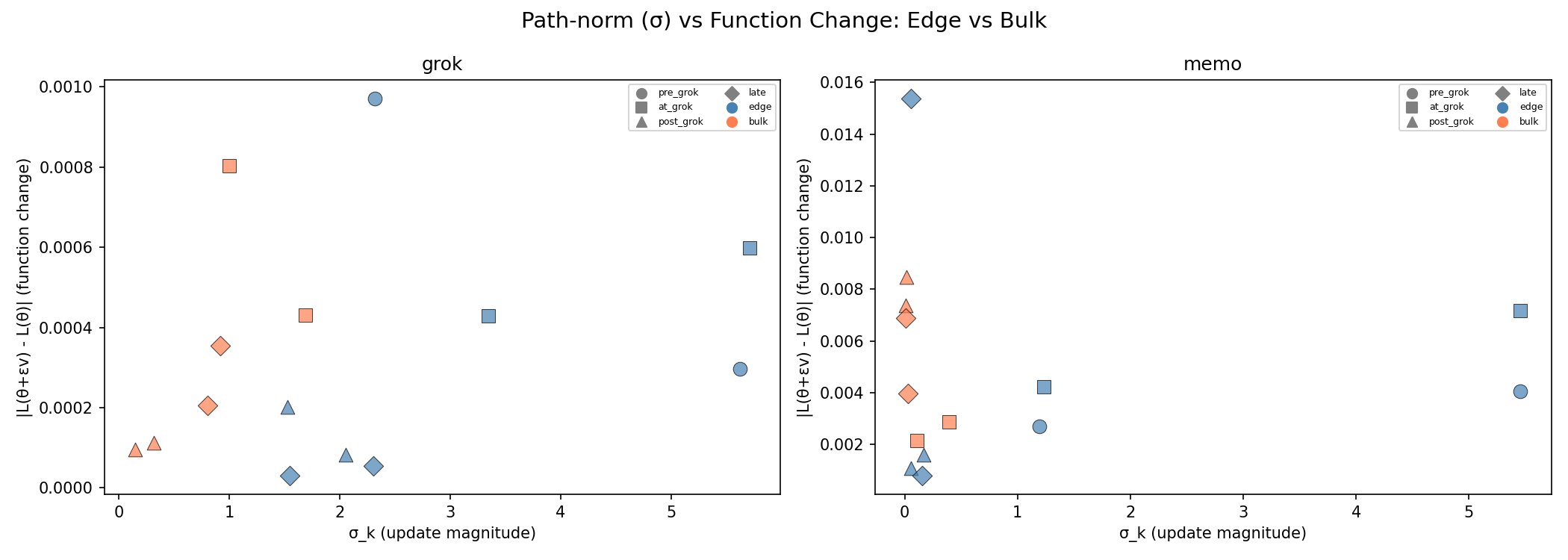}
    \caption{Path-norm ($\sigma_k$ = update magnitude) vs.\ function change ($|\Delta L|$) per direction.
    \textbf{Grokked} (left): large $\sigma$ with near-zero function change---compression (weight space motion without functional consequence).
    \textbf{Memorized} (right): small $\sigma$ with large function change---fragility (any motion disrupts the function).}
    \label{fig:pathnorm}
\end{figure}

\Cref{fig:pathnorm} makes this concrete: the grokked edge has $\sigma = 2.3$ but $|\Delta L| = 0.00005$ (large motion, zero functional change); the memorized bulk has $\sigma = 0.01$ but $|\Delta L| = 0.015$ (tiny motion, large disruption).

\subsection{Resolution: Orientation Matters, Displacement Doesn't}

The paradox resolves through a geometric distinction articulated by the thesis's Davis-Kahan stability bound: the spectral gap protects the \emph{orientation} of $v_1$ (perturbation $\sin\theta \leq \norm{\Delta G}_F / g$, and $g$ is large), while the low curvature $h_1 = 0.08$ means the loss is flat \emph{along} $v_1$.

\begin{quote}
The model's function depends on \emph{where} $v_1$ points (removing it is catastrophic).\\
It does not depend on \emph{motion along} $v_1$ (perturbing it is harmless).\\
Weight decay selects flat directions that preserve the learned function while reducing parameter norm.
\end{quote}

This reframes weight decay: rather than ``regularizing'' in a generic sense, WD specifically selects directions along which the loss landscape is flat---directions where parameters can be compressed without functional cost.
The spectral edge \emph{is} this selected direction.

% ══════════════════════════════════════════════════════════════════════
\section{Nonlinear Re-Encoding, Not Information Loss}
\label{sec:nonlinear}
% ══════════════════════════════════════════════════════════════════════

\begin{figure}[t]
    \centering
    \includegraphics[width=0.7\textwidth]{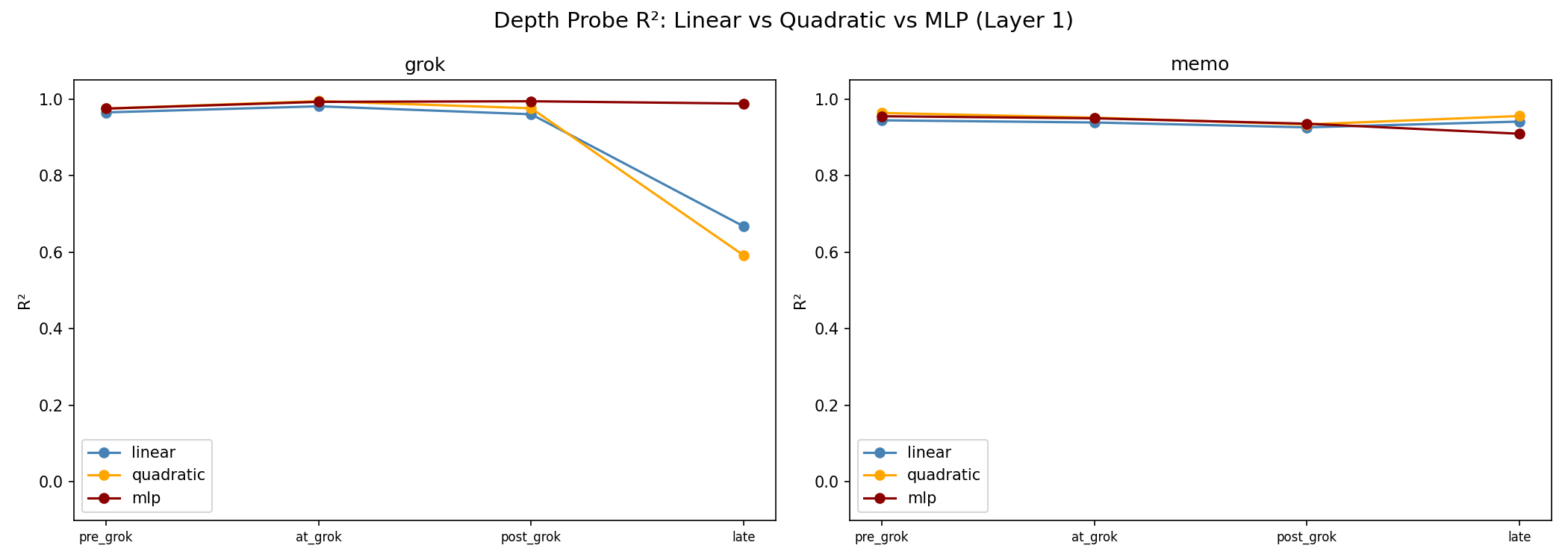}
    \caption{Probe $R^2$ for depth from layer-1 representations (Dyck).
    \textbf{Left (grokked)}: linear probes degrade late ($0.97 \to 0.67$); quadratic probes fail even worse ($0.59$); but MLP probes hold at $0.99$ throughout.
    \textbf{Right (memorized)}: all probes remain high.
    Depth information is not lost---it is nonlinearly re-encoded by weight-decay compression.}
    \label{fig:probes}
\end{figure}

A linear probe for depth on layer-1 representations drops from $R^2 = 0.98$ at grokking to $R^2 = 0.86 \pm 0.01$ late (3 seeds).
This has been interpreted as information loss.
It is not.

A shallow MLP probe (one hidden layer, 64 units) recovers $R^2 = 0.990 \pm 0.003$ on the same representations (\Cref{fig:probes}).
Quadratic probes also fail ($R^2 = 0.59$), so the re-encoding is genuinely deep nonlinear, not polynomial.

The information is always present; weight decay changes how it is \emph{encoded}, compressing linearly accessible representations into a more abstract form.
This is consistent with the ablation paradox: the model knows depth (MLP $R^2 = 0.99$) but doesn't store it in a linearly readable direction (linear $R^2 = 0.86$, dropping as low as $0.67$ for individual seeds at late training).

% ══════════════════════════════════════════════════════════════════════
\section{Weight Decay Intervention: The Causal Test}
\label{sec:wd_intervention}
% ══════════════════════════════════════════════════════════════════════

\begin{figure}[t]
    \centering
    \includegraphics[width=\textwidth]{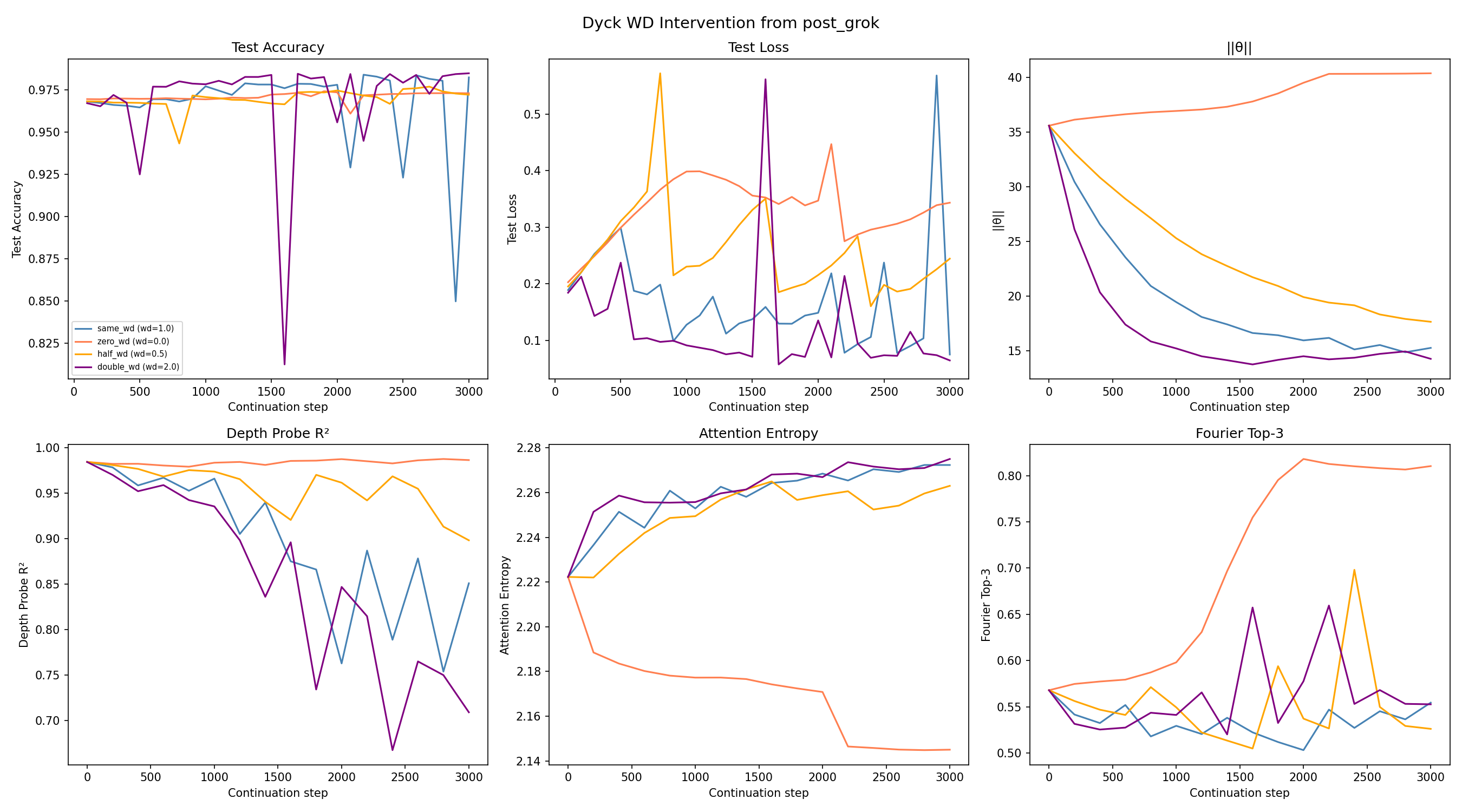}
    \caption{Continuing training from a post-grok checkpoint under four WD conditions (Dyck).
    Removing WD ($\omega = 0$, red): accuracy holds at 0.97, probe $R^2$ recovers to 0.99, entropy drops, parameter norm balloons.
    Doubling WD ($\omega = 2$, purple): accuracy rises slightly, probe $R^2$ drops further, entropy maximizes.
    WD drives compression of the representation; the learned algorithm is independent of WD.}
    \label{fig:wd_intervention}
\end{figure}

We continue training from a post-grok checkpoint with four weight-decay conditions (\Cref{fig:wd_intervention}).

\begin{table}[t]
\centering
\caption{Weight decay intervention (post-grok, Dyck).  WD drives compression (lower $R^2$, higher entropy, lower $\norm{\theta}$) but not generalization (accuracy survives removal).}
\label{tab:wd}
\begin{tabular}{lcccc}
\toprule
\textbf{Condition} & \textbf{Test acc} & \textbf{Linear $R^2$} & \textbf{Entropy} & $\norm{\theta}$ \\
\midrule
$\omega = 0$ (remove) & 0.973 & \textbf{0.987} & 2.15 & 40.4 \\
$\omega = 0.5$        & 0.972 & 0.898 & 2.26 & 17.7 \\
$\omega = 1.0$ (same) & 0.982 & 0.851 & 2.27 & 15.3 \\
$\omega = 2.0$ (double) & \textbf{0.985} & 0.709 & \textbf{2.28} & \textbf{14.3} \\
\bottomrule
\end{tabular}
\end{table}

Three conclusions:
\begin{enumerate}[nosep]
    \item \textbf{The algorithm survives without WD.}  Accuracy drops only 0.009 when WD is removed entirely.  The generalizing computation was learned at grokking and persists without continued compression.
    \item \textbf{WD drives the nonlinear re-encoding.}  Removing WD restores linear $R^2$ from 0.85 to 0.99.  Doubling WD pushes it to 0.71.  MLP $R^2$ is 0.99 throughout.  The encoding changes; the information does not.
    \item \textbf{WD drives uniform attention.}  Entropy increases monotonically with $\omega$ ($2.15 \to 2.28$).  Without WD, attention becomes less uniform---the counting algorithm relaxes toward position-specific patterns.
\end{enumerate}

This establishes the causal direction: weight decay is the engine of post-grok compression, not the engine of generalization.
The spectral edge's post-grok behavior (widening gap, growing $\sigma_1/\sigma_2$) is driven by WD compression along $v_1$, not by continued learning.

% ══════════════════════════════════════════════════════════════════════
\section{Three Universality Classes of the Spectral Edge}
\label{sec:universality}
% ══════════════════════════════════════════════════════════════════════

The spectral edge does not behave identically across tasks.
We identify three regimes, predicted by the balance of gradient driving (Term~3) and WD damping (Term~2) in the gap flow~\eqref{eq:gap_flow}:

\begin{table}[t]
\centering
\caption{Three universality classes of spectral edge behavior.  The class is determined by the relative strength of gradient driving (Term~3) vs.\ WD damping (Term~2) post-grok.}
\label{tab:classes}
\begin{tabular}{llcccc}
\toprule
\textbf{Class} & \textbf{Task} & \textbf{Grad\% (late)} & \textbf{Rotation} & \textbf{Func.\ $R^2$} & \textbf{Character} \\
\midrule
\textsc{Functional}  & Mod-arith & high & moderate & high (single $\omega$) & edge carries Fourier modes \\
\textsc{Mixed}        & Dyck      & 87\% & 18$^\circ$ & 0.80 & partial functional content \\
\textsc{Compression}  & SCAN      & 2\%  & 8$^\circ$  & 0.04 & functionally empty \\
\bottomrule
\end{tabular}
\end{table}

\paragraph{Functional edge (modular arithmetic).}
Term~3 remains active post-grok: the gradient continues to project onto $v_1$, maintaining functional content.
Edge directions carry distinct Fourier frequencies ($\omega = 25$ for addition, $\omega = 29$ for multiplication in the discrete-log basis~\citep{xu2026functional_modes}).
Edge/bulk separation is clear: the edge concentrates on task-relevant modes while the bulk is diffuse.

\paragraph{Mixed edge (Dyck).}
Term~3 partially persists (87\% gradient late), keeping $v_1$ functionally active ($R^2 = 0.80$).
The edge rotates 18$^\circ$ per window---more than SCAN's frozen axis, less than the bulk's 52$^\circ$.
Fourier analysis in the depth basis shows $5.2\times$ concentration above uniform at grokking, but no edge/bulk separation (all directions project onto the single eigenmode: depth counting).

\paragraph{Compression edge (SCAN).}
Term~2 dominates almost completely (99.8\% WD at grok, 98\% late).
The edge freezes (8$^\circ$ rotation) and empties functionally ($R^2 = 0.04$).
It becomes a pure compression axis---a direction along which WD reduces $\norm{\theta}$ without affecting the function.
The model has enough redundancy (1.5M params for a 2048-sample task) that WD finds ample flat directions.

\paragraph{Prediction.}
The universality class is determined by $\text{capacity} / \text{task complexity}$.
High redundancy $\to$ compression edge (WD finds flat directions easily).
Low redundancy $\to$ functional or mixed edge (WD must share directions with the gradient).
For large language models, we predict the compression class will dominate, with the spectral edge becoming a parameter-reduction axis rather than a feature-learning one.

% ══════════════════════════════════════════════════════════════════════
\section{What the Models Learn: Fourier Structure}
\label{sec:fourier}
% ══════════════════════════════════════════════════════════════════════

The spectral edge describes \emph{how} training dynamics select computations.
Fourier analysis describes \emph{what} computations are selected.
These are related but distinct: the edge is a property of the training trajectory; the Fourier structure is a property of the learned model.

\subsection{Dyck: A Dual Frequency Structure}

\begin{figure}[t]
    \centering
    \includegraphics[width=\textwidth]{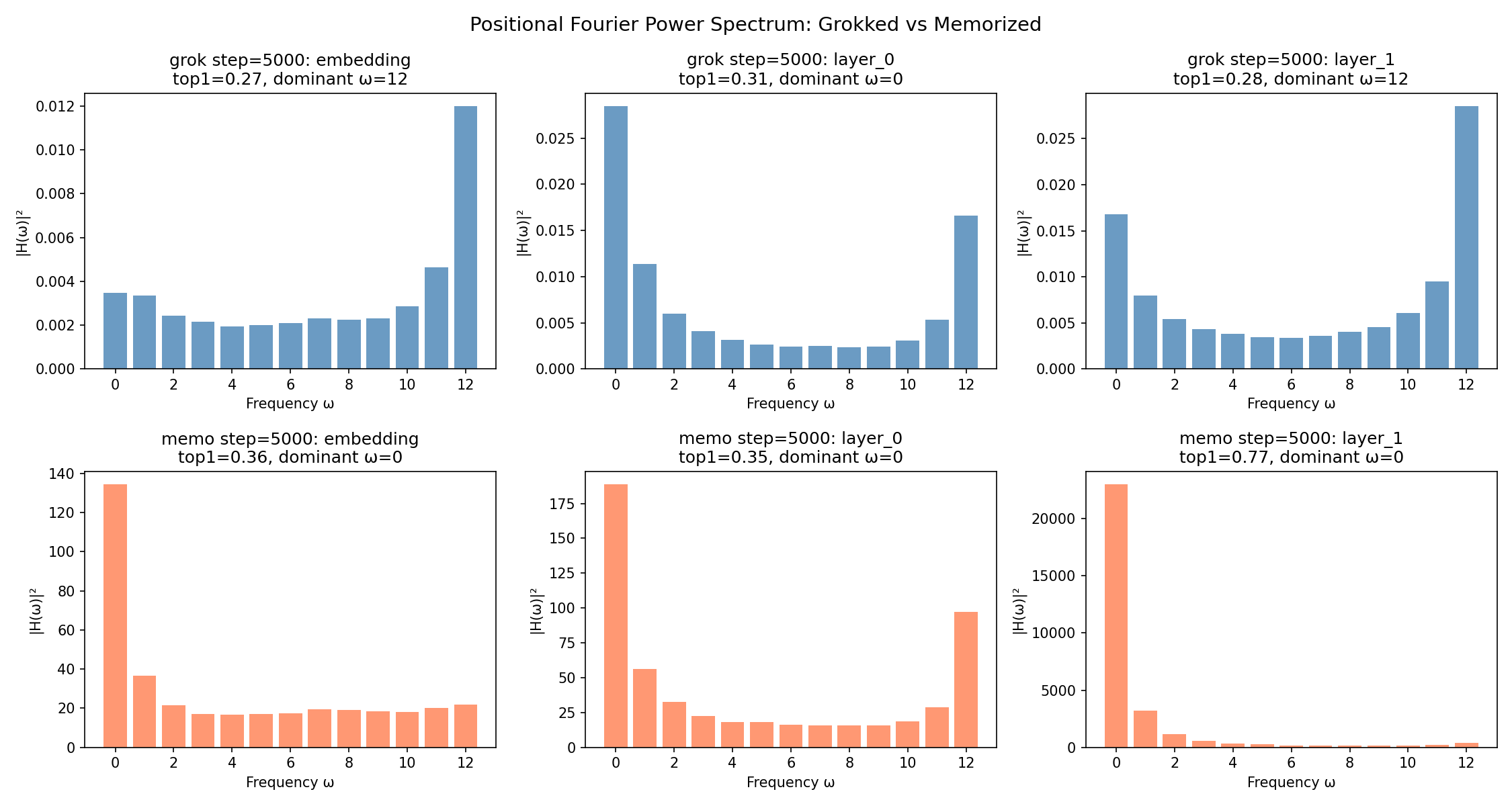}
    \caption{Positional Fourier power spectra (Dyck, post-training).
    \textbf{Grokked} (top): energy peaks at $\omega = 12$ (Nyquist for $T = 24$, matching open/close alternation).
    \textbf{Memorized} (bottom): 77\% of layer-1 energy collapses to $\omega = 0$ (DC).
    The grokked model encodes token identity in the frequency domain; the memorized model stores a position-invariant template.}
    \label{fig:power_spectrum}
\end{figure}

Hidden representations of the grokked model concentrate at $\omega = 12$---the Nyquist frequency for sequences of length $T = 24$, which corresponds to the binary alternation between open and close tokens (\Cref{fig:power_spectrum}).
The memorized model collapses to $\omega = 0$ (the DC mode): all positions have approximately the same representation.

\begin{figure}[t]
    \centering
    \includegraphics[width=\textwidth]{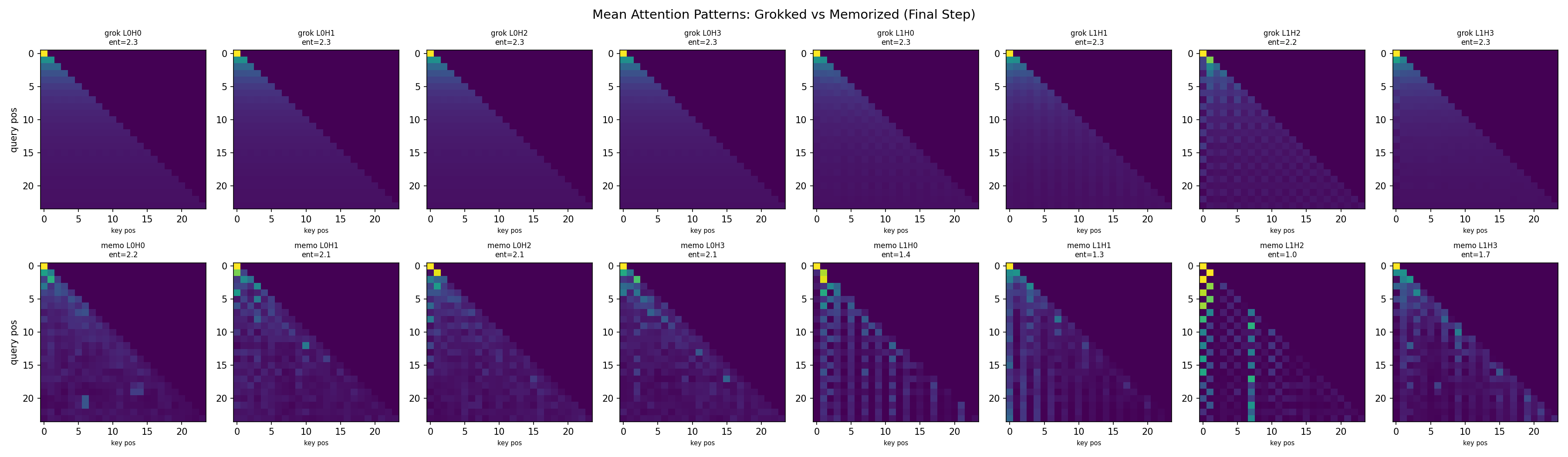}
    \caption{Mean attention patterns (Dyck).
    \textbf{Grokked} (top): all layer-0 heads converge to uniform backward attention (smooth gradient = the counting algorithm).
    \textbf{Memorized} (bottom): peaked, position-specific patterns (bright spots = a lookup table).}
    \label{fig:attention}
\end{figure}

Attention patterns tell the complementary story (\Cref{fig:attention}): all four layer-0 heads of the grokked model converge to \emph{near-perfect uniform backward attention} (KL from uniform $< 0.001$, entropy = 2.28).
Uniform attention computes an unweighted average of past token embeddings, which linearly encodes the fraction of open tokens seen so far---i.e., the depth.
This is the counting algorithm, and it is the $\omega = 0$ mode of the attention pattern.

The grokked Dyck model thus has a dual Fourier structure:
\begin{itemize}[nosep]
    \item \textbf{Representations}: $\omega = 12$ (``what to count''---token identity).
    \item \textbf{Attention}: $\omega = 0$ (``how to count''---uniform averaging).
\end{itemize}

This parallels modular arithmetic, where the learned computation aligns with a single group character~\citep{nanda2023grokking, liu2023understanding}.
The difference is that Dyck's ``group'' is the binary token set $\{+1, -1\}$ with cumulative summation, not $\mathbb{Z}_p$ with modular addition.
The Fourier structure is diagnostic when symmetry exists, but the deeper invariant---that grokking selects a low-dimensional algorithmic computation---holds regardless.

\subsection{Compositional Factorization}

\begin{figure}[t]
    \centering
    \includegraphics[width=\textwidth]{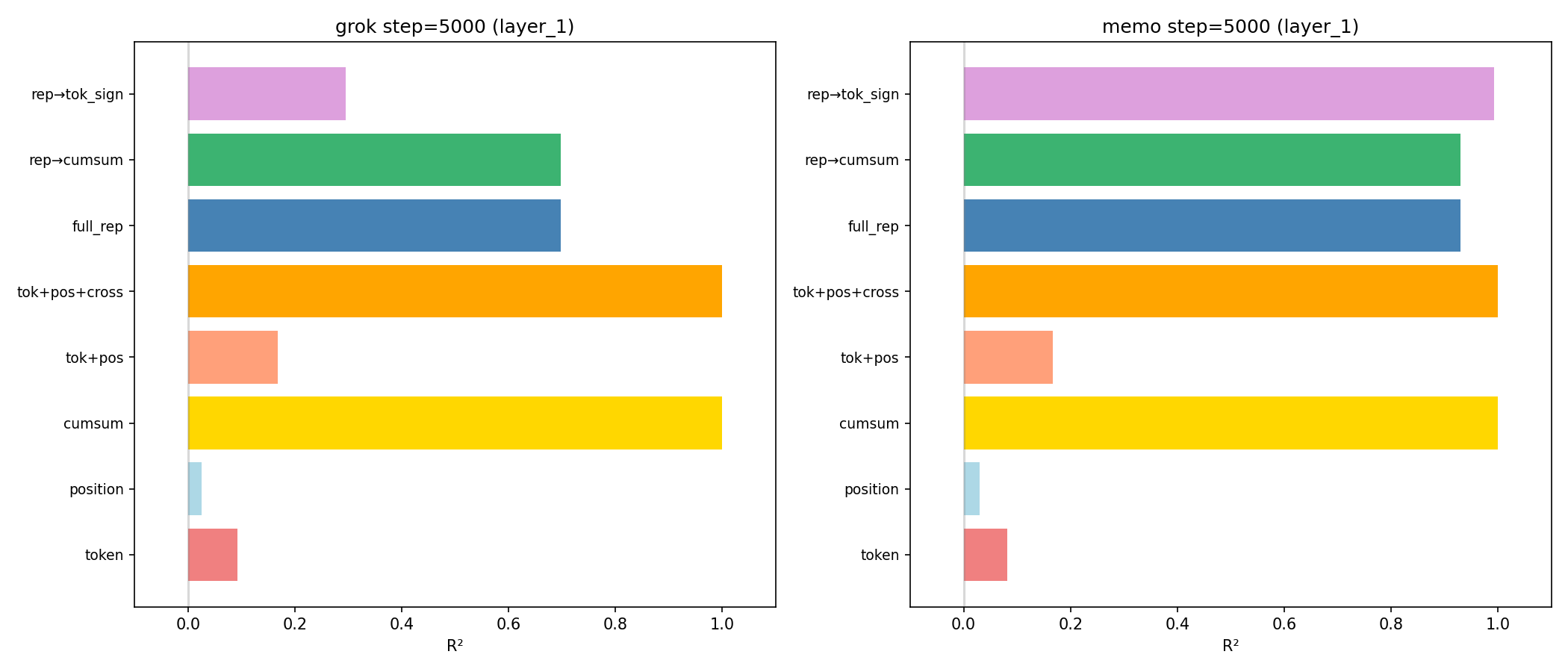}
    \caption{Compositional probing (Dyck, layer 1).
    Cross-term features (token $\times$ cumsum) achieve $R^2 = 1.0$ for both models, confirming depth is compositionally determined.
    But the grokked model does not linearly encode token identity ($R^2 = 0.30$ vs.\ $0.99$ for memorized)---it abstracts away from surface features.}
    \label{fig:composition}
\end{figure}

Cross-term features (token type $\times$ running sum) achieve $R^2 = 1.0$ for both models (\Cref{fig:composition}), confirming that depth is compositionally determined.
But the grokked model does not linearly store the components: token identity $R^2 = 0.30$ (vs.\ 0.99 for memorized), running depth $R^2 = 0.70$ (vs.\ 0.93).
The grokked model computes the composition without explicitly representing its factors---a compressed, abstract encoding.

\subsection{Depth Representation Geometry}

\begin{figure}[t]
    \centering
    \includegraphics[width=\textwidth]{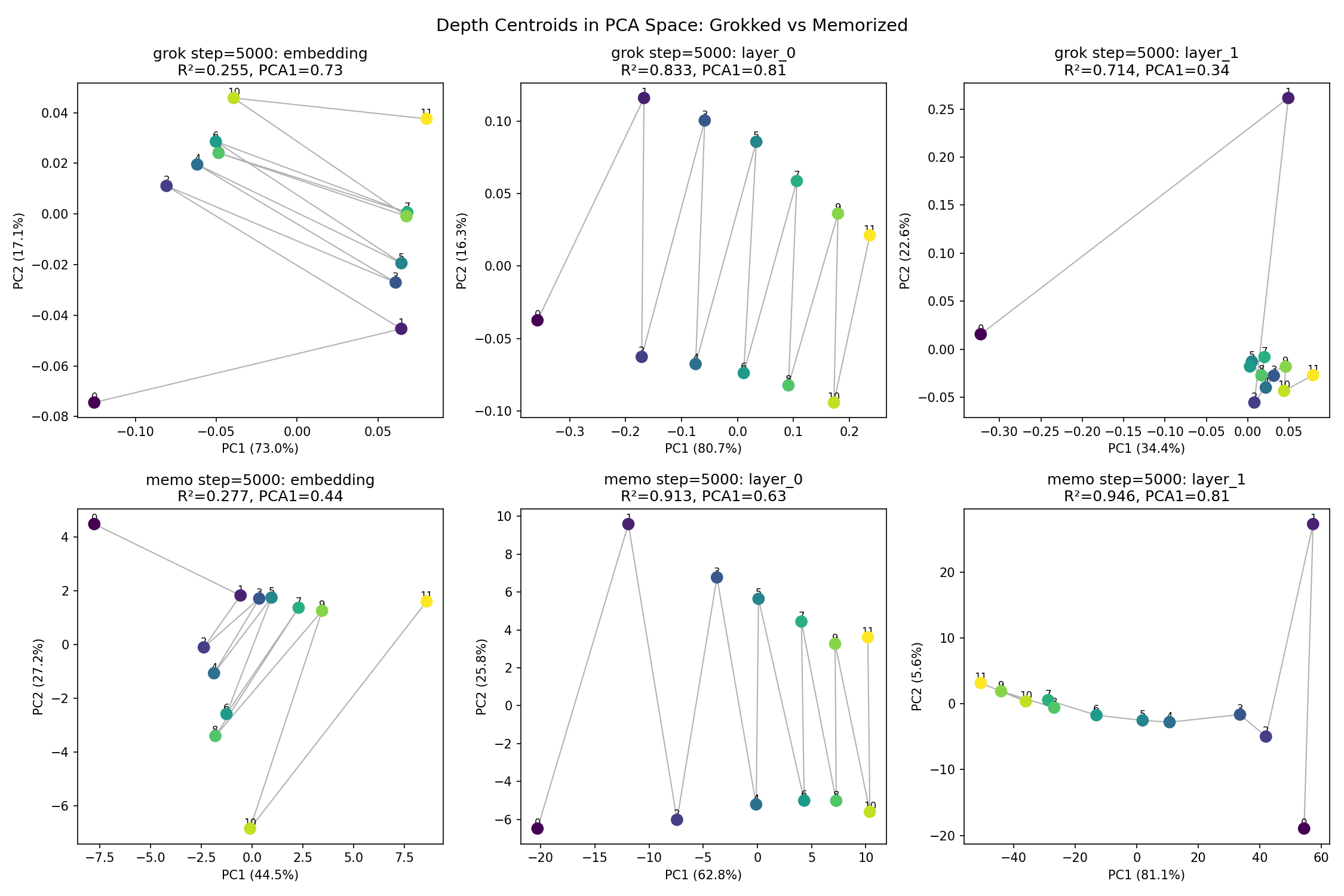}
    \caption{Depth centroids in PCA space (layer 1).
    \textbf{Grokked} (top): distributed across dimensions (PCA$_1 = 34\%$), smooth arc.
    \textbf{Memorized} (bottom): stretched along one axis (PCA$_1 = 81\%$), $150\times$ larger inter-depth distances.
    Higher linear $R^2$ in the memorized model (0.95 vs.\ 0.71) is a signature of explicit storage, not better computation.}
    \label{fig:depth_geometry}
\end{figure}

A surprising inversion: the memorized model has \emph{higher} linear probe $R^2$ for depth (0.95 vs.\ 0.71) and $150\times$ larger inter-depth distances (\Cref{fig:depth_geometry}).
It encodes depth in a single high-variance direction (PCA$_1 = 81\%$); the grokked model distributes it across multiple dimensions (PCA$_1 = 34\%$).
This is not a defect---it is the representation-space signature of the ablation paradox: the grokked model's depth encoding is compressed and non-linear, requiring an MLP to read out ($R^2_\text{MLP} = 0.99$).

\subsection{Fourier Analysis of the Spectral Edge}

\begin{figure}[t]
    \centering
    \begin{subfigure}[t]{0.48\textwidth}
        \includegraphics[width=\textwidth]{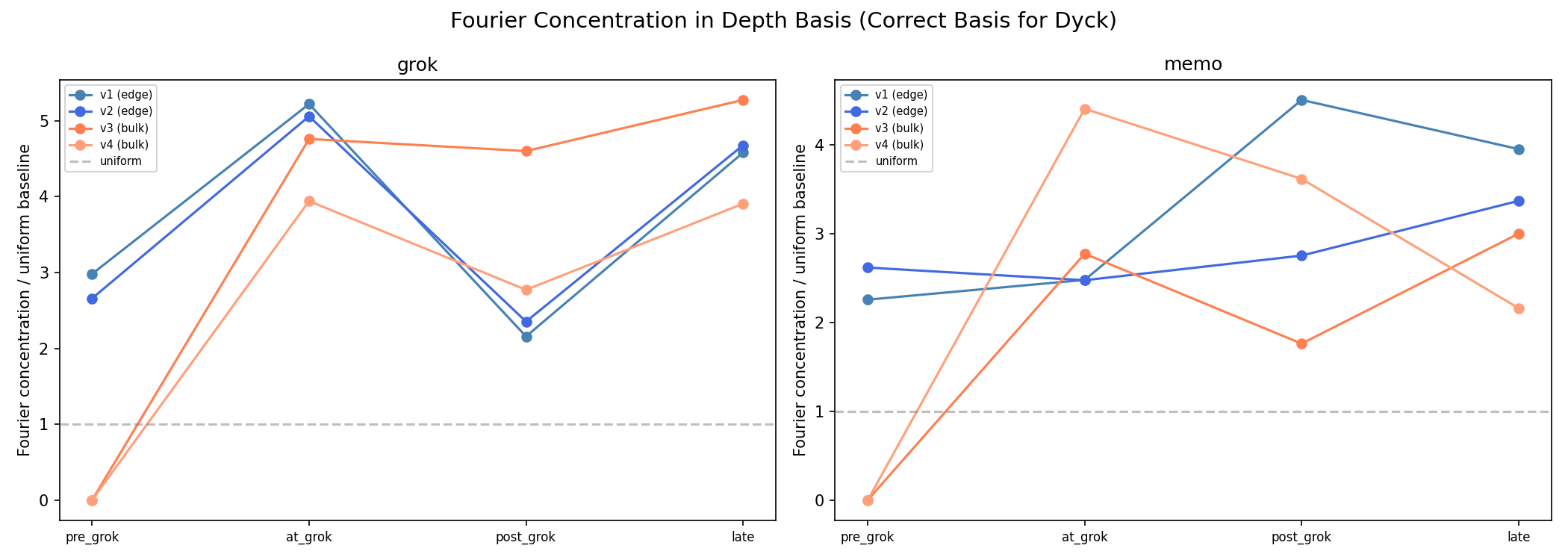}
        \caption{Fourier concentration in the depth basis. At grokking, the edge reaches $5.2\times$ uniform.}
    \end{subfigure}
    \hfill
    \begin{subfigure}[t]{0.48\textwidth}
        \includegraphics[width=\textwidth]{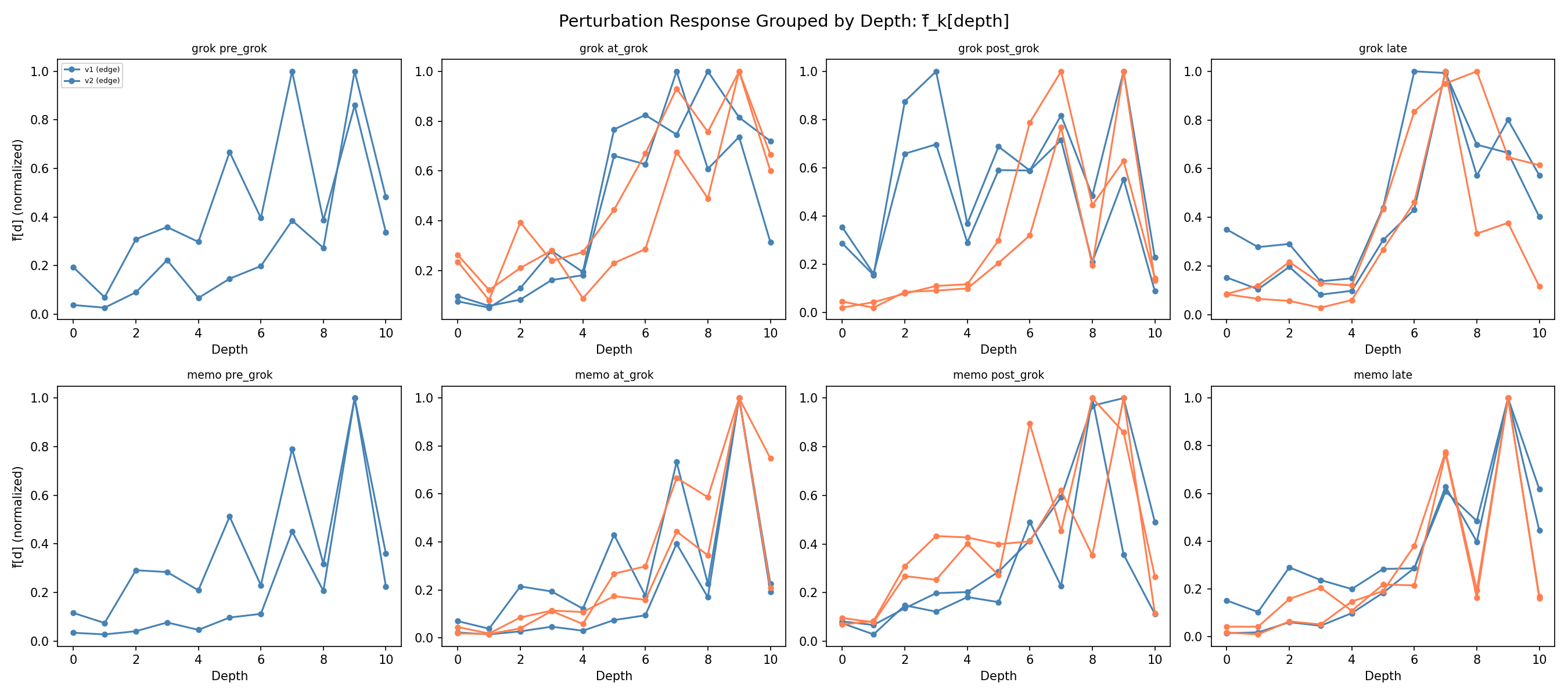}
        \caption{Perturbation response $\bar{f}_k[d]$ grouped by depth: structured variation (deep vs.\ shallow).}
    \end{subfigure}
    \caption{Fourier analysis of Gram edge directions in the depth basis (the correct functional coordinate for Dyck).}
    \label{fig:depth_fourier}
\end{figure}

For the spectral edge directions themselves, we compute the perturbation response $f_k(x) = \norm{\Delta h_k(x)}^2$, group by depth $d \in \{0, \ldots, 12\}$, and apply the DFT---the same protocol as the companion paper~\citep{xu2026functional_modes} but using depth rather than $(a+b) \bmod p$ as the grouping variable.

At grokking, the edge achieves Fourier concentration $F = 0.87$ ($5.2\times$ above uniform, \Cref{fig:depth_fourier}), peaking at $\omega = 1$---a single oscillation across depth levels (shallow positions respond differently from deep ones).
This parallels the mod-arith result ($19\times$ for addition at $\omega = 25$), though the absolute elevation is lower because the domain is smaller (13 depth levels vs.\ 97 residues).

Unlike modular arithmetic, edge and bulk directions do not separate in the depth basis: both concentrate at $\omega = 1$.
This is because Dyck has one functional eigenmode (depth counting), and all Gram directions project onto it.
The edge/bulk distinction becomes quantitative (update magnitude) rather than qualitative (functional content) when there is only one mode---consistent with the ``mixed edge'' universality class.

% ══════════════════════════════════════════════════════════════════════
\section{Replication and Controls}
\label{sec:controls}
% ══════════════════════════════════════════════════════════════════════

\begin{figure}[t]
    \centering
    \begin{subfigure}[t]{0.48\textwidth}
        \includegraphics[width=\textwidth]{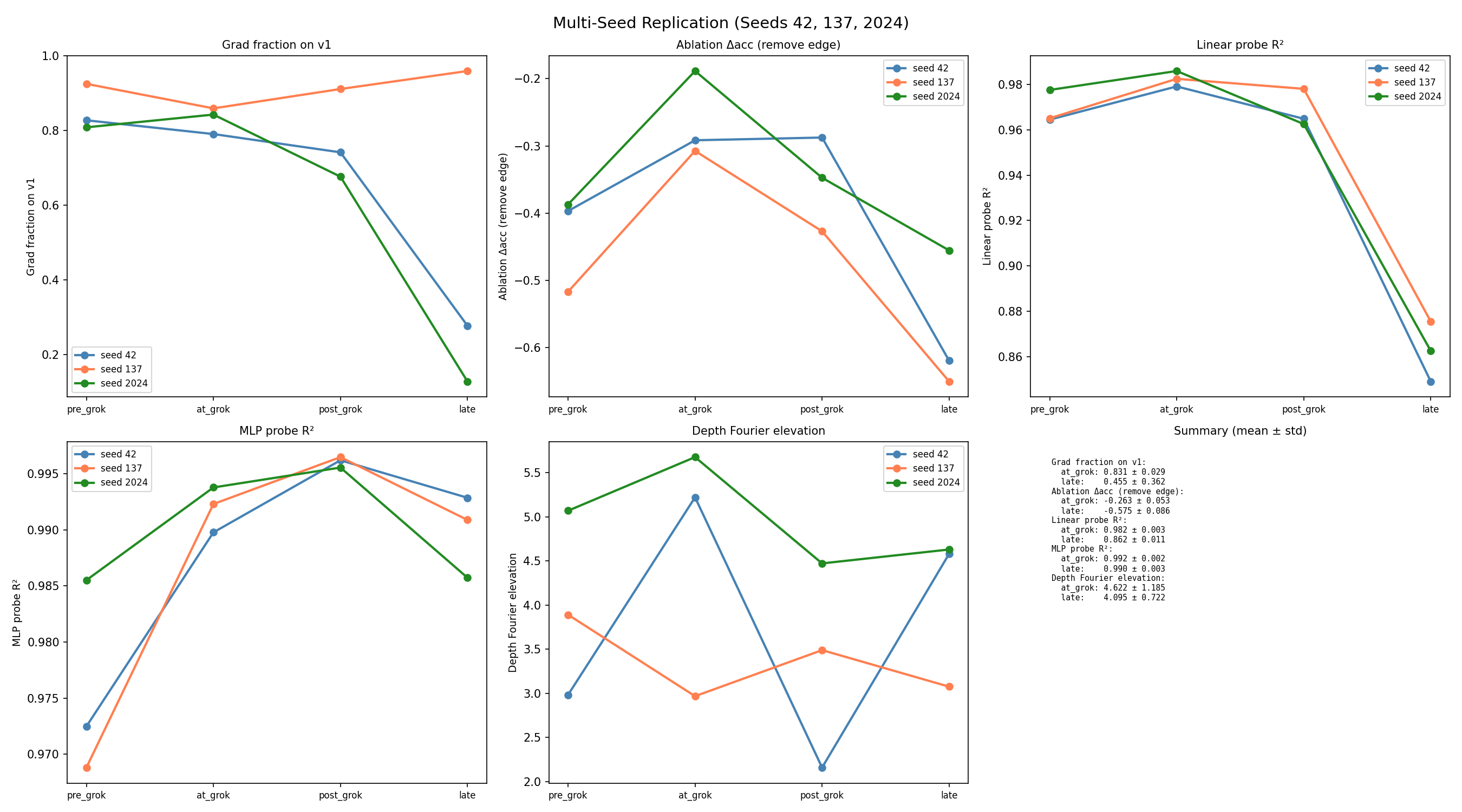}
        \caption{Multi-seed replication (Dyck, 3 seeds).}
        \label{fig:multiseed}
    \end{subfigure}
    \hfill
    \begin{subfigure}[t]{0.48\textwidth}
        \includegraphics[width=\textwidth]{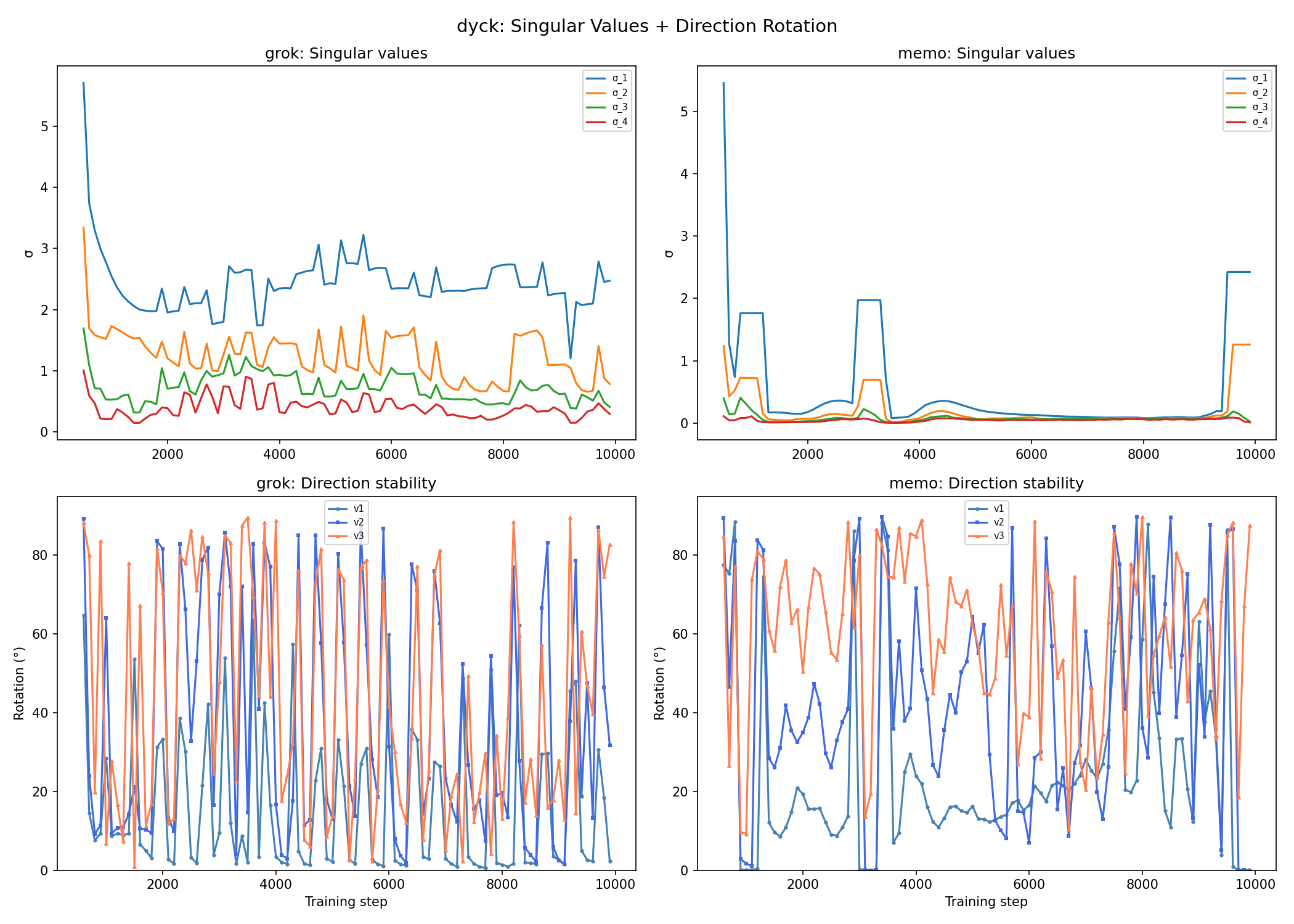}
        \caption{Singular values and direction rotation (Dyck, grok vs.\ memo).}
        \label{fig:rotation}
    \end{subfigure}
    \caption{Robustness. (a)~All key findings replicate across seeds 42, 137, 2024. MLP probe $R^2 = 0.990 \pm 0.003$ is the most robust.
    (b)~The stability hierarchy: $v_1$ rotates $18^\circ$ (moderate), $v_3$ rotates $52^\circ$ (unstable). SCAN's $v_1$ rotates only $8^\circ$ (frozen).}
    \label{fig:controls}
\end{figure}

\paragraph{Multi-seed.}
All findings replicate across seeds 42, 137, 2024 (\Cref{fig:multiseed}):
ablation $\Delta\text{acc} = -0.58 \pm 0.09$;
linear $R^2 = 0.86 \pm 0.01$;
MLP $R^2 = 0.990 \pm 0.003$;
depth Fourier elevation $= 4.1 \pm 0.7\times$.

\paragraph{Rotation stability.}
The thesis predicts $\alpha_\text{dom} > \alpha_\text{gap} > \alpha_\text{sub}$.
We confirm: $\alpha_1 = 0.77 > \alpha_2 = 0.16$ at grokking; $v_1$ rotates $18^\circ$, $v_3$ rotates $52^\circ$ (\Cref{fig:rotation}).
SCAN's edge is even more frozen ($8^\circ$), consistent with its compression-edge character.

\paragraph{Random controls.}
Ablating random directions has zero effect (20 trials/phase, all $|\Delta\text{acc}| < 10^{-4}$).
The edge is $>$4000$\times$ more impactful (\Cref{fig:ablation}).

\paragraph{SCAN full suite.}
All results extend to SCAN: edge ablation $\Delta\text{acc} = -0.49$ (grok) vs.\ $-0.03$ (memo); Hessian curvature $0.001$ (grok edge) vs.\ $1.83$ (memo bulk); random ablation has zero effect ($>10^7\times$ edge/random ratio).

% ══════════════════════════════════════════════════════════════════════
\section{Discussion}
\label{sec:discussion}
% ══════════════════════════════════════════════════════════════════════

\subsection{One Mechanism, Multiple Projections}

The central mechanism---grad-WD alignment at grokking producing a two-phase edge---manifests differently depending on where you look:

\begin{center}
\begin{tabular}{lll}
\toprule
\textbf{Space} & \textbf{Pre-grok} & \textbf{Post-grok} \\
\midrule
Weight space & Gradient-driven $v_1$ & WD-driven $v_1$ \\
Function space & Perturbation changes output & Perturbation is flat \\
Representation space & Depth linearly accessible & Depth nonlinearly encoded \\
Attention space & Position-specific (lookup) & Uniform (counting algorithm) \\
Fourier (representations) & Distributed & $\omega = 12$ (token structure) \\
Fourier (edge, depth basis) & Moderate concentration & $5.2\times$ elevation at $\omega = 1$ \\
\bottomrule
\end{tabular}
\end{center}

These are not independent findings---they are projections of the same event.
The grad-WD alignment produces a flat compression direction (weight space), which preserves the function while compressing parameters (function space), which drives nonlinear re-encoding of depth (representation space), which coincides with the crystallization of the counting algorithm (attention space).

\subsection{What Weight Decay Actually Does}

Our results suggest a specific reframing:

\begin{quote}
Weight decay selects flat directions in the loss landscape---directions along which parameters can be compressed without functional cost.
The spectral edge \emph{is} this selected direction.
Post-grok, WD drives the model along the edge, reducing $\norm{\theta}$ while maintaining $f(\theta)$.
The resulting compression nonlinearly re-encodes representations, making them more abstract and less linearly accessible.
\end{quote}

This connects to the flat minima literature~\citep{hochreiter1997flat, cohen2021gradient}: WD biases toward solutions where the loss landscape is flat, and the spectral edge is the direction of maximal flatness.
It also connects to implicit bias and margin maximization~\citep{lyu2020gradient, lyu2024dichotomy}: the edge is the direction of margin maximization projected onto the update subspace.
The observation that training dynamics concentrate in a tiny subspace~\citep{gurari2018gradient, sagun2017empirical} is the starting point; we show that this subspace has internal structure (the edge/bulk distinction) with a dynamical lifecycle.

\subsection{A Practical Prediction: WD Scheduling}

Since WD drives compression (lower linear $R^2$) without affecting the algorithm (accuracy stable), a concrete prediction follows:

\begin{quote}
\emph{Train with high WD to grok, then reduce WD to obtain linearly-accessible representations without losing the algorithm.}
\end{quote}

Our intervention shows this works: removing WD post-grok recovers $R^2_\text{linear}$ from 0.85 to 0.99 while keeping accuracy at 0.97.
This could be useful for interpretability: a model trained with WD scheduling would have a readable representation of its learned computation, without sacrificing generalization.

\subsection{Limitations}

Both tasks are small-scale (150K--1.5M parameters, 50--2048 training samples).
The thesis~\citep{xu2026spectral_edge} confirms $k^* \leq 3$ and gap-loss correlation for models up to 124M parameters, but the grad-WD decomposition has not been tested at that scale.
The depth basis for Dyck is ``correct'' by construction; for tasks where the natural functional coordinate is unknown, the framework requires discovering the basis---an open problem.
Our correlational evidence (alignment iff grokking) does not prove that alignment \emph{causes} grokking; it could be a downstream consequence of a deeper mechanism.

% ══════════════════════════════════════════════════════════════════════
% References
% ══════════════════════════════════════════════════════════════════════

\bibliographystyle{plainnat}

\end{document}